\documentclass[10pt,twocolumn,letterpaper]{article}
\pdfoutput=1
\usepackage{wacv}
\usepackage{times}
\usepackage{epsfig}
\usepackage{graphicx}
\usepackage{amsmath}
\usepackage{amssymb}
\usepackage{booktabs}
\usepackage{tabularx}

\newcommand{\abs}[1]{\lvert#1\rvert}

\def\wacvPaperID{****} 

\wacvapplicationstrack 

\wacvfinalcopy 

\ifwacvfinal
\usepackage[breaklinks=true,bookmarks=false]{hyperref}
\else
\usepackage[pagebackref=true,breaklinks=true,colorlinks,bookmarks=false]{hyperref}
\fi

\begin{document}

\title{DNN Filter for Bias Reduction in Distribution-to-Distribution Scan Matching}

\author{Matt McDermott\\
Tufts University\\
Medford MA, USA\\
{\tt\small matthew.mcdermott@tufts.edu}
\and
Jason Rife\\
Tufts University\\
Medford MA, USA\\
{\tt\small jason.rife@tufts.edu}
}

\maketitle
\thispagestyle{empty}

\begin{abstract}
Distribution-to-distribution (D2D) point cloud registration techniques such as the Normal Distributions Transform (NDT) can align point clouds sampled from unstructured scenes and provide accurate bounds of their own solution error covariance-- an important feature for safety-of-life navigation tasks. 
D2D methods rely on the assumption of a static scene and are therefore susceptible to bias from range-shadowing, 
self-occlusion,
moving objects, and distortion artifacts 
as the recording device moves between frames.
Deep Learning-based approaches can achieve higher accuracy in dynamic scenes by relaxing these constraints, however, DNNs produce uninterpretable solutions which can be problematic from a safety perspective. 
In this paper, we propose a method of down-sampling LIDAR point clouds to exclude voxels that violate the assumption of a static scene and introduce error to the D2D scan matching process. 
Our approach uses a solution consistency filter-- identifying and suppressing voxels where D2D contributions disagree with local estimates from a PointNet-based registration network. 
Our results show that this technique provides significant benefits in registration accuracy, and is particularly useful in scenes containing dense foliage.

\end{abstract}


\begin{figure}[h]
\centering
\includegraphics[width=3.4in]{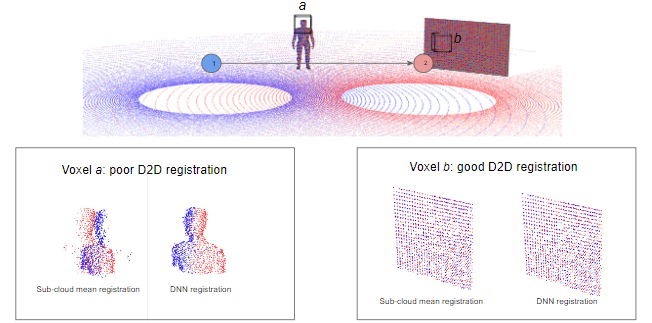}
\caption{Comparison of registration of points from two scans captured at locations 1 and 2, contained within voxels \textit{a} and \textit{b}. The DNN registration estimate disagrees with the D2D estimate for the registration of the head, so the points in voxel \textit{a} will be ignored.}
\label{fig:meanFittingBad}
\end{figure}


\section{Introduction}

Our goal in this paper is to to demonstrate a technique for identifying and discarding voxels that introduce error in Distribution-to-Distribution point cloud registration. In order to discuss our proposed method we must first introduce the current state of the art in point cloud registration, provide insight into some of the challenges associated with this task, and discuss how they relate to our proposed approach.

Distribution-to-Distribution methods represent clusters of points as probability density functions and use Maximum Likelihood Estimation to find the transformation that best aligns the two PDFs \cite{Biber, D2DNDT}. The accuracy of a LIDAR point cloud registration is a function of both the quality of the sensor and the geometry of the scene in which the data is recorded.
Analytical techniques permit the direct calculation of error bounds for solution covariance by considering the cost function at convergence \cite{censi, D2DNDT} or directly via a least-squares analysis of the PDF occupying each voxel \cite{2DICET}. While these registration techniques boast impressive accuracy prediction on static scenes, motion of the ego-vehicle can introduce systemic bias through range-shadowing \cite{sphericalICET}, self-occlusion \cite{behindTheCurtain}, and ''rolling shutter'' motion distortion \cite{KITTI_distortion}. For scan registration algorithms to be useful in real-world applications, they must maintain robustness in the presence of these forms of bias.

Deep Learning methods can achieve superior accuracy to D2D techniques by relaxing some of their assumptions and forfeiting solution interpratability.
Deep Neural Network (DNN) based point cloud registration can be broadly divided between correspondence-free methods, in which a DNN directly estimates a transformation in an end-to-end manner, and correspondence-based methods, which alternate between identifying corresponding features in each scan and minimizing distances between correspondences \cite{DNNsurvey} similar to the early \textit{Iterative Closest Point} (ICP) algorithm \cite{ICP}.
Correspondence-based methods can use a DNN to identify keypoints in each frame before estimating correspondences analytically \cite{dmlo} or can identify correspondences using an additional network \cite{deepvcp}. 
Generally, correspondence-based methods achieve higher accuracy registering larger point clouds (such as full LIDAR scans) while correspondence-free methods excel on clouds sampled from individual objects or small portions of a LIDAR scan \cite{DNNsurvey}. One common approach is to subdivide large LIDAR scans into voxels and apply a single end-to-end network to each cell simultaneously as a batch operation \cite{PointNetLk, voxelnet}. Some batch methods also aggregate local estimates to gain additional contextual information about the scene \cite{pointnetvlad}. 


Partial overlap between point clouds, which occurs in regions subject to range-shadowing and self-occlusion, presents a problem for registration. 
Range-shadowing tends to inject systemic bias across a large swath of voxels in a constant direction normal to the edge of each shadow \cite{sphericalICET}, while perspective shift bias is highly localized on the surfaces of objects that appear different from various angles \cite{behindTheCurtain}.
Geometric strategies exist to reduce the bias of range-shadowing in D2D algorithms, either through ground plane removal \cite{lego, GP-ICP} or through a spherical coordinate representation \cite{KITTI_distortion, sphericalICET, hassani2021new}, however, error introduced by self-occlusion is significantly more complex.  
One approach to mitigate self-occlusion is to train a network to fill in occluded regions of point clouds before registration is attempted.
In \textit{MaskNet}, the authors directly estimate inlier points from one scan using candidate points from the other \cite{masknet}. Self-supervised methods overcome partial overlap by jointly learning geometric representations, keypoint detection, and correspondence \cite{prnet, NDT_transformer}. 
\textit{BTCdet} augments preliminary estimates from a backbone network by filling in points on occluded surfaces using a network trained \textit{a priori} on commonly encountered object shapes \cite{behindTheCurtain}.


Although there is a significant body of literature on both Machine Learning-based and Distribution-to-Distribution point cloud registration methods, there is limited work in the context of leveraging ML to improve distribution-based approaches.
Tang et al. use Gaussian Process Regression to reduce bias in NDT by directly estimating a correction factor to add to their preliminary registration \cite{biascorrection}. While their strategy is capable of reducing registration error, allowing an uninterpretable model to directly adjust odometry estimates negates much of the safety benefits of NDT. Our proposed solution differs from this form of mixed method in that the uninterpretable component solely exists to down-sample data that is likely to introduce bias to the geometric registration routine. 
Crucially, the reduction of total information due to the down-sampled point cloud can be accounted for in the optimization routine, automatically adjusting the solution error covariance prediction component of the registration algorithm as problematic voxels are removed from a scan.

Random Sample Consensus (RANSAC) is a general statistical tool for outlier removal \cite{RANSAC}, and has been used for ground plane segmentation in NDT \cite{RANSAC_ground_plane}. Kanhere and Gao also use measurement consensus to improve computational efficiency in NDT, and to devise a metric for quantifying localization reliability \cite{kanhere2019lidar}. Other approaches assume the magnitude of residuals at a converged solution will be normally distributed and simply discard any voxels that produce residuals more than a few standard deviations away from the mean \cite{magnussonThesis}. Unfortunately, classical outlier routines rely on a large plurality of voxels in a scene being well-conditioned. This can lead to these approaches excluding voxels with accurate solution contributions if they are outnumbered by voxels with consistently biased estimates. The density of the points within a voxel is inversely proportional to the distance of the voxel from the origin, so the regions of a scan that are most susceptible to perspective shift bias (close to the origin) are also on average the most populated and heavily weighted. 

Moving objects within the frame also present additional challenges for scan registration. A fine voxel gird will automatically discount small objects moving at high speed if they do not occupy the same cell at the converged solution. 
While our strategy of considering per-voxel registration to identify bias from occlusion does not directly suppress moving objects, it allows the rejection threshold of our moving object detection routine to be raised to to the point that its only function is to discard obviously moving objects. 


In section 2 we provide more detail on the detection theory underpinning our filtering process. In section 3 we describe our network structure and training procedure. In sections 4 and 5 we introduce and share the results of two experiments to validate our method; a detection analysis where we demonstrate how our approach is used to identify and exclude biased voxels, and a demonstration on various KITTI datasets where we show how our filter preforms on real LIDAR data. In section 6 we provide further discussion into our results. Lastly, in section 7 we summarize our findings and provide insights for future work.

\section{Filtering}

Our approach is to test for bias in an estimate without having access to the true solution, introducing a secondary estimation technique and considering the difference between the two. If there is agreement between the two uncorrelated estimates, it is unlikely that either is significantly flawed. If the two estimates differ by a sufficient magnitude, however, one of the two registration techniques is likely biased and the points inside the offending voxel should be ignored.
As long as the errors between D2D and DNN remain uncorrelated, pruning voxels where the two methods disagree provides the flexibility of a learned registration scheme while retaining statistical robustness.

Because the translation errors from DNN and D2D solutions in the $x$, $y$, and $z$ components are all independent and normally distributed (as we will demonstrate in the following section), the overall magnitude of error in the translation vector for each method relative to the ground truth forms a chi-square distribution with three degrees of freedom. This relationship is described by equation (\ref{eq:ChiSquare1}), where $x$, $y$, and $z$ represent the true translation solutions about each axis and $\hat{x}$, $\hat{y}$, and $\hat{z}$ represent the associated estimates output by each technique.

\begin{equation}\label{eq:ChiSquare1}
    Q_{3} \sim (x - \hat{x})^2 + (y - \hat{y})^2 + (z - \hat{z})^2
\end{equation}


When constructing our monitor statistic, however, we require a single metric to summarize the spread of residuals between the DNN and D2D solutions. It is convenient to difference the two measurements in vector space before unifying solution components into a single chi-square distribution
as the sum of two normally distributed random variables is also normal, while calculating the difference of noncentral chi-square distributions is nontrivial. 
Here, the vector difference between D2D and DNN estimates (and consequently, the difference between their errors) is represented as $\Delta$. The distribution of these errors for a given voxel registration is then defined as $Q_{3, \Delta}$:

\begin{equation}\label{delta}
    \begin{split}
    \mathbf{E}_{DNN} = \mathbf{X}_{DNN} - \mathbf{X}_{true} \\
    \mathbf{E}_{D2D} = \mathbf{X}_{D2D} - \mathbf{X}_{true} \\
    \Delta = \mathbf{E}_{D2D} - \mathbf{E}_{D2D} \\
    Q_{3, \Delta} = | \Delta | 
    \end{split}
\end{equation}

\begin{figure}[h]
\centering
\includegraphics[width=3.4in]{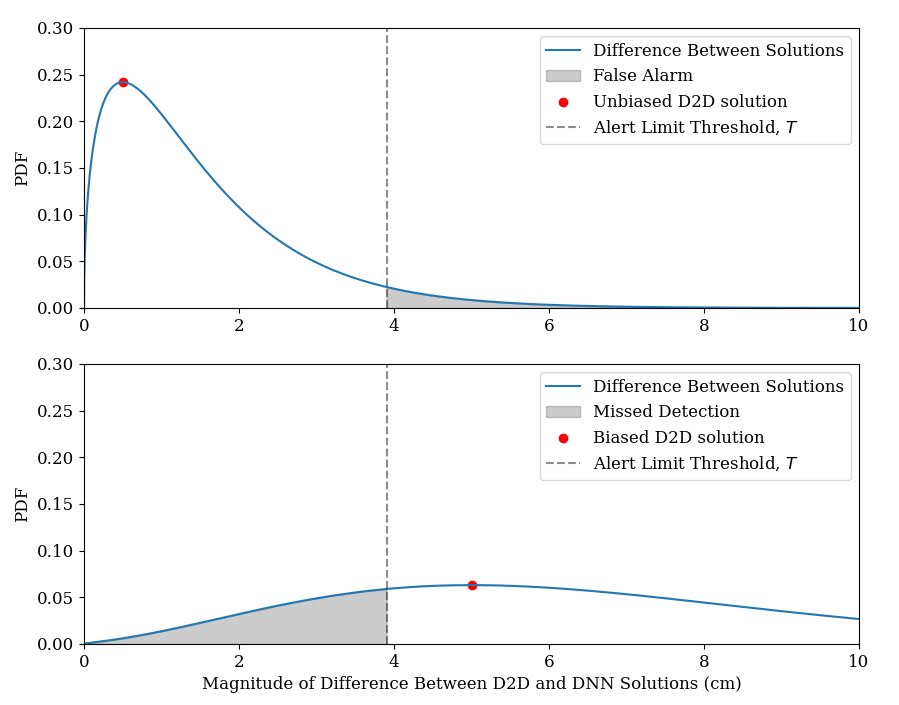}
\caption{Trade-off between false alarms and missed detections}
\label{fig:falseAlarm}
\end{figure}

Figure (\ref{fig:falseAlarm}) displays the distribution of $Q_{3,\Delta}$ for two cases. The top plot demonstrates the spread of $Q_{3,\Delta}$ when neither the D2D or DNN approach is systemically biased for a particular test sample (such as the wall inside voxel \textit{b} in Fig. (1)). The bottom plot in Fig. (\ref{fig:falseAlarm}) illustrates an alternate case of $Q_{3, \Delta}$ on a test sample that produces significant systemic bias in the D2D solution, as would be the case in the presence of heavy shadowing or self-occlusion such as the partially occluded human model inside voxel \textit{a} in Fig. (1). In Fig. (\ref{fig:falseAlarm}), the shaded region within each plot represents the areas in which the monitor statistic fails to correctly characterize the behavior of the system. This demonstrates the trade-off between false alarms and missed detection when selecting a threshold for the alert limit. The alert limit in figure (\ref{fig:falseAlarm}) is set to produce a 5\% chance of false alarm (shown in the top plot) where the DNN incorrectly excludes an unbiased D2D solution estimate. At this threshold, the lower plot represents the associated risk of missed detection for test sample where D2D registration is biased by 5cm. In this example, the selected threshold value produces a significantly higher level of false negatives than false positives. 
Framing outlier rejection as a regression problem, rather than one of binary classification, allows rejection criteria to be tuned after training depending on the desired characteristics of the system.

For both cases in Fig. (\ref{fig:falseAlarm}), it is apparent that relying on differences between two solutions misses a large portion of the useful information. If the geometic approach was prone to uniformly distributed errors for any scene, there would be no reason to employ an alternate registration technique (and thus lose useful info). As is discussed in \cite{sphericalICET, behindTheCurtain}, however, certain geometric properties in a scene violate the assumption of a static scene significantly enough to introduce systemic bias. This highlights the inherent tradeoff between solution accuracy and solution integrity. It may be acceptable for a noisy filter to reject a large portion of the voxels in a well-conditioned scene, thereby slightly reducing average registration accuracy, if the filter is also likely to reject significant bias in other scenes that would otherwise prevent convergence altogether.



\section{DNN Structure}


Our network is constructed as a variant of \textit{Iterative PCR-Net} \cite{sarode2019pcrnet}, which leverages \textit{PointNet} \cite{PointNet} encoding to directly estimate the transformation that best registers two point clouds. Here, input point clouds are concatenated and passed through feed-forward layers with shared weights before a Max-Pooling symmetric function is applied to produce a feature vector, which is then passed through a series of simple feedforward layers. 

\begin{figure}[h]
\centering
\includegraphics[width=3.3in]{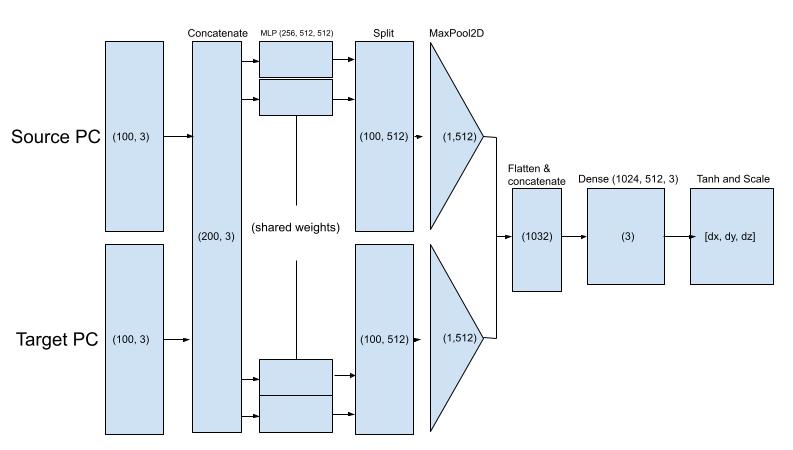}
\caption{DNN structure}
\label{fig:DNN}
\end{figure}


The purpose of our network is to either validate or reject the per-voxel D2D solutions, so we limit the network output to a similar solution space and only estimate translation within each voxel. 
Removing rotation from the solution vector prevents the network from having to contend with any forms of rotational ambiguity (i.e. when attempting to register an isolated cylinder) that may detract from the training process. 
Linearizing each voxel to a translation-only estimate is consistent with other registration algorithms such as the \textit{Lucas \& Kanade} (LK) algorithm \cite{LK}, and its application to the point cloud registration task with a global PointNet in \textit{PointNetLK} \cite{PointNetLk}.
A side benefit of applying our network on a per-voxel basis is that any out of distribution behavior is compartmentalized to the offending region of the frame. 

Like other local point cloud registration techniques, an initial estimate within the basin of attraction is required to begin the scan matching process. This can be achieved using an estimate from another sensor, such as wheel odometry, a dynamic vehicle model, or simply by zeroing out initial transformation and attempting a coarse-to-fine approach with large voxels.
In our implementation, a preliminary solution estimate is first achieved by allowing D2D registration to converge before sampling points to feed to the DNN.  



\section{Experiments}

We validate our method of bias mitigation in distribution-based scan matching through two experiments

\subsection{Per-voxel accuracy}

\begin{figure}[h]
\centering
\includegraphics[width=2.0in]{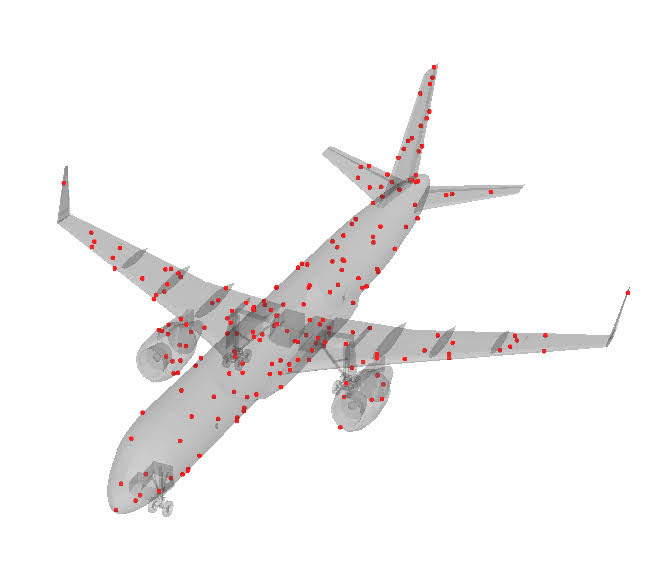}

\includegraphics[width=0.8in]{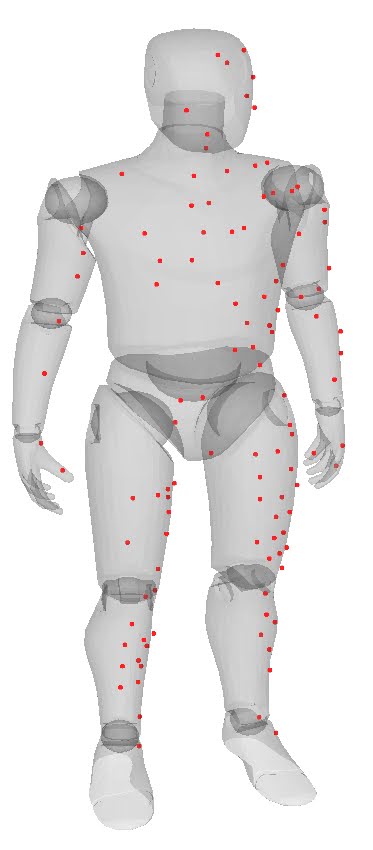}

\includegraphics[width=2.0in]{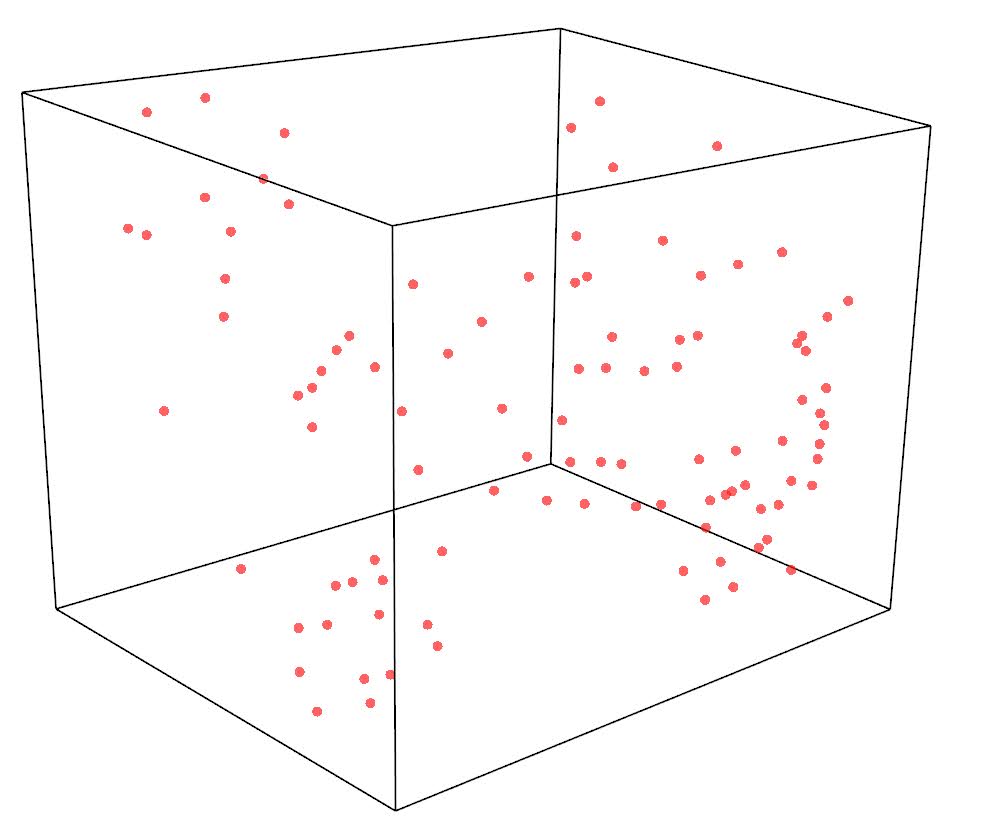}
\caption{100 point keyframe clouds for uniformly sampled \textit{ModelNet40} object (top), simulated LIDAR scan recorded from from \textit{ModelNet40} object (middle), and voxelized KITTI (bottom)}
\label{fig:experiment1clouds}
\end{figure}

Our first experiment seeks to quantify the differences in registration accuracy between the two approaches.
Here, we compare solutions from our PointNet against the distance between the means of each sub-cloud, which is the basis for the D2D registration contributions for each voxel.
The first two trials of this experiment test simple point clouds generated from objects in the popular \textit{ModelNet40} dataset \cite{modelnet40}, and represent cases where each voxel completely encapsulates a single object with no ground plane or external-occlusion present. The first is an unrealistic case where clouds are formed by uniformly sampling points on the surface of each object. The second repeats this process, but generates more realistic point clouds from the \textit{ModelNet40} objects by randomly transforming each model and simulating LIDAR scans from the visible surfaces. This test provides a middle ground between the unrealistic point clouds in the first trial and real world data in the following trial. 

In the third trial we consider the per-voxel registration accuracy of sub-clouds sampled from voxels in the \textit{KITTI} dataset. Here, we gather training data from each scan using spherical voxels as described by \cite{sphericalICET} to mitigate the effects of range-shadowing. Eliminating range-shadows in the training data greatly reduces the need to share information between cells. For all cases we train the network on point clouds containing 100 points from each scan per voxel, a value tuned to align with the optimal minimum number of points required for D2D registration \cite{sphericalICET, GP-ICP}.
Training on small sections of larger point clouds requires additional considerations, which are discussed in detail in the following section. 

\subsection{Improving Total Error with Filter}

\begin{figure}[h]
\centering
\includegraphics[width=3.0in]{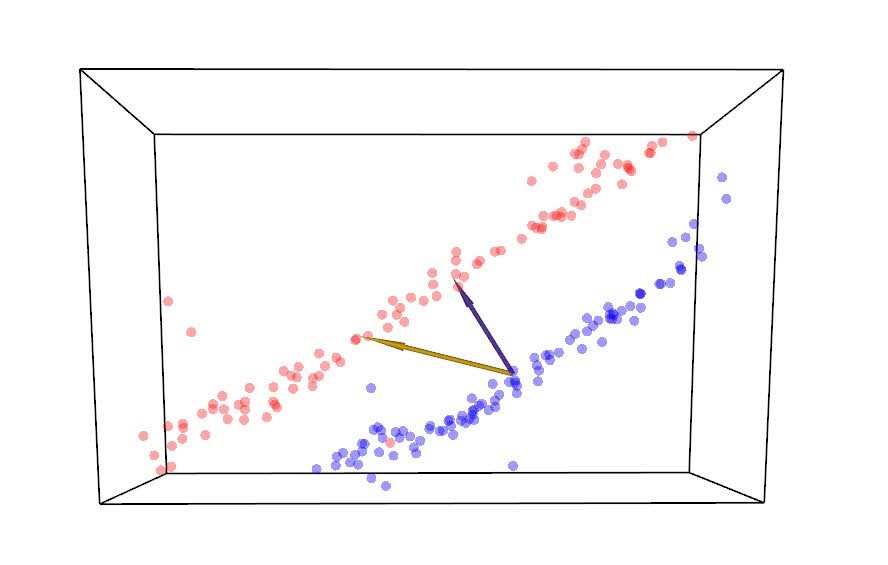}
\caption{Overhead view of a voxel containing partially partially overlapping sections of the same wall captured in two LIDAR scans. The true translation to align the clouds is drawn in yellow, however, there is insufficient information to constrain solutions in the direction aligned with the surface of the wall. The reduced-dimension ground truth obtained using using Eq. (\ref{eq:DNNpruning}) is shown in purple.  }
\label{fig:compactProjection}
\end{figure}

Our second experiment makes use of real LIDAR data, again taken from the KITTI benchmark dataset. Here, LIDAR data contains shadowing, occlusion, and non-returns. Surfaces in this test also extend beyond the limits of each cell. Estimating local registration solutions with no consideration of the larger context can hinder training, as the information present within a cell is often insufficient to properly estimate translation \cite{2DICET, shimojo1989occlusion}. Generally, the spread of points in the direction normal to a surface is randomly distributed due to sensor noise, while the spread of points tangent to the surface tends to be more uniform and structural, following the underlying shape of the object, rather than the properties of the sensor.
Since distributions are cut arbitrarily short in surface-aligned directions by the voxel grid, attempting to align centers of these "overly extended" directions provides no useful information, and in fact, can negatively impact solution accuracy \cite{2DICET}. 
For that reason, we integrate a process of ambiguous axis suppression into our loss function when training the registration network using equation (\ref{eq:DNNpruning}).
As shown in Fig. (\ref{fig:compactProjection}), the our filter is also constructed to only count components of residuals normal to compact directions of each scan. 

\begin{equation}\label{eq:D2Dpruning}
    {}^{(j)} \textbf{x}_{D2D} = {}^{(j)}\textbf{L}\: {}^{(j)}\textbf{U}^T ({}^{(j)}\Bar{\textbf{x}}_r - {}^{(j)}\bar{\textbf{x}}_n)
\end{equation}

Equation (\ref{eq:D2Dpruning}) describes the process of projecting the vector representing the reference and new scan means $\bar{\mathbf{x}}_r$ and $\bar{\mathbf{x}}_n$, respectively for each voxel $j$ into a reduced dimension residual vector. 
Here, $\mathbf{U}$ is the rotation matrix (obtained via eigendecomposition of the covariance of the points in voxel $j$) that aligns the principal axis of distribution $J$ relative to the world frame, and $\mathbf{L}$ is an truncation matrix.

\begin{equation}\label{eq:DNNpruning}
    {}^{(j)} \textbf{x}_{DNN} = {}^{(j)}\textbf{L}\: {}^{(j)}\textbf{U}^T {}^{(j)}\hat{\textbf{x}}_{DNN}
\end{equation}

Equation (\ref{eq:DNNpruning}) represents the process through which extended components of the residual vector estimated by the DNN are excluded. Here, ${}^{(j)}\hat{\textbf{x}}_{DNN}$ represents the raw translation estimate output by the network for voxel $j$.
It is important to note how this process also makes use of the same $\mathbf{U}$ and $\mathbf{L}$ matrices as equation (\ref{eq:D2Dpruning}). This means that the shape of the keyframe distribution (that has already been calculated in the classical distribution-to-distribution routine) is used to inform the pruning matrix which components of the DNN residual are important to consider, because they likely represent a true offset between surfaces, and which components are merely due to a structural spread of points. The total offset, ${}^{(j)}\Delta\mathbf{x}$, between useful components of the D2D and DNN solution is then calculated as follows

\begin{equation}
    {}^{(j)}\Delta\mathbf{x} = \abs{{}^{(j)}\mathbf{x}_{DNN} - {}^{(j)}\mathbf{x}_{D2D}}
\end{equation}

If ${}^{(j)}\Delta\mathbf{x}$ is larger than a threshold $T$, all points located inside voxel $j$ are excluded and the associated contributions to the overall solution vector from voxel $j$ are ignored. When constructing our DNN, we achieved best results by applying the projection and dimension reduction matrices inside the loss function, rather than as distinct input parameters passed to the network. This allows us to train using data from both KITTI, as well as simulated LIDAR scans of objects in the ModelNet40 dataset. 

In Experiment II we validate our strategy on three KITTI datasets. The first contains 160 sequential LIDAR point clouds recorded from a vehicle moving between 3 and 7 $\frac{m}{s}$ in a busy urban setting. The second dataset contains 194 sequential frames where the vehicle moves between 11 and 22 $\frac{m}{s}$ on a highway through a dense forest. The third dataset contains 800 sequential LIDAR frames recorded while moving through a forested residential neighborhood at 8 $\frac{m}{s}$.




\section{Results}

\begin{table}[h]
    \caption{Experiment I: Per Voxel RMS Translation Error (cm)} 
    \setlength{\tabcolsep}{0.3\tabcolsep}
    \begin{tabularx}{0.48\textwidth} {  
      | >{\centering\arraybackslash}X
      | >{\centering\arraybackslash}X
      | >{\centering\arraybackslash}X 
      | >{\centering\arraybackslash}X
      | >{\centering\arraybackslash}X
      | >{\centering\arraybackslash}X
      | >{\centering\arraybackslash}X
      | >{\centering\arraybackslash}X
      | >{\centering\arraybackslash}X | }
    \hline
     &  DNN (ours) & D2D \\
    \hline
    ModelNet40 ~\small{(uniformly sampled)} & 5.48 & 3.61 \\
    \hline
    ModelNet40 (shadowed) & 5.82 & 15.29  \\
    \hline
    KITTI & 9.79 & 14.04 \\ 
    \hline
    \end{tabularx}
    \label{tab:experiment1}
\end{table}




\begin{figure*}[h]
\centering
\includegraphics[width=1.7in]{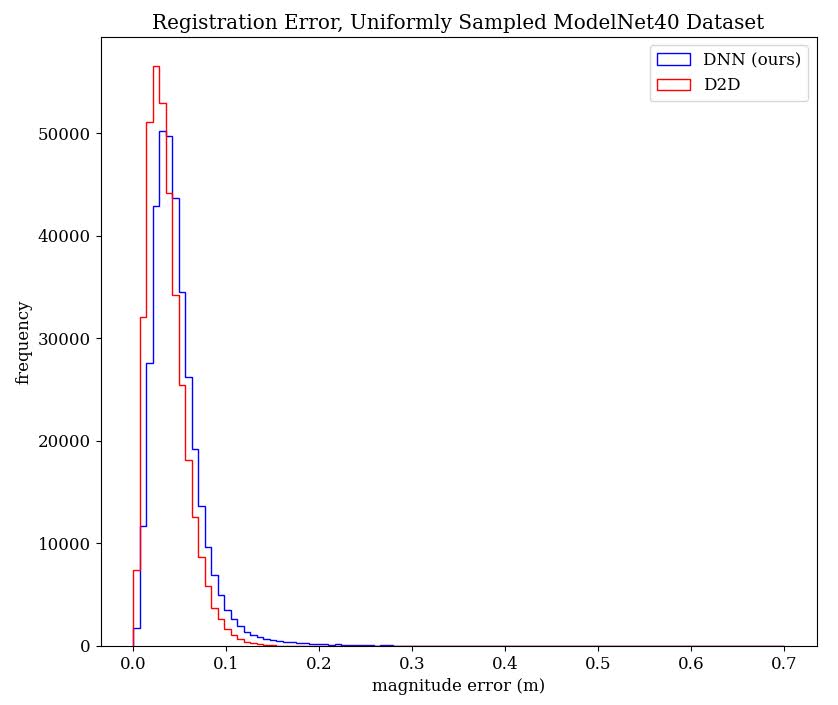}
\includegraphics[width=1.6in]{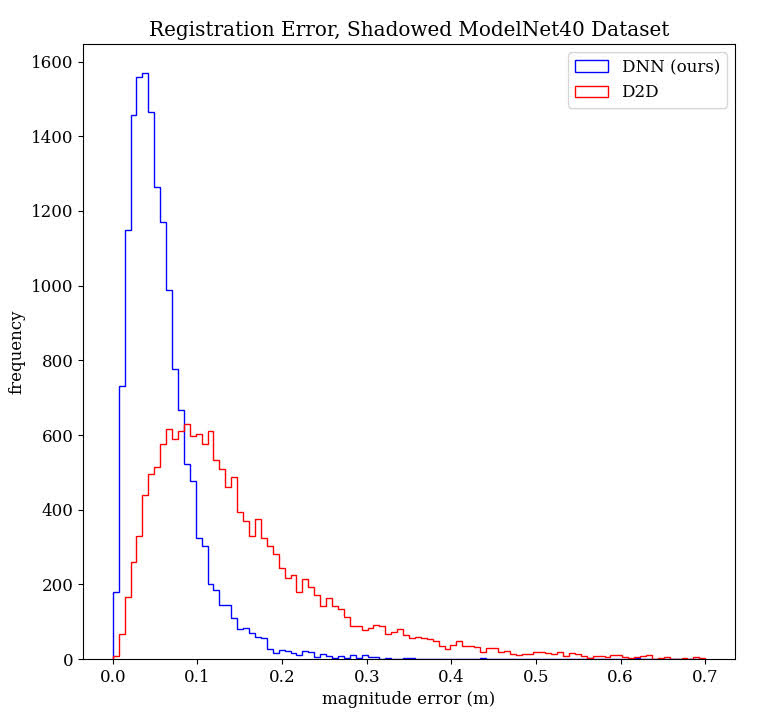}
\includegraphics[width=2.in]{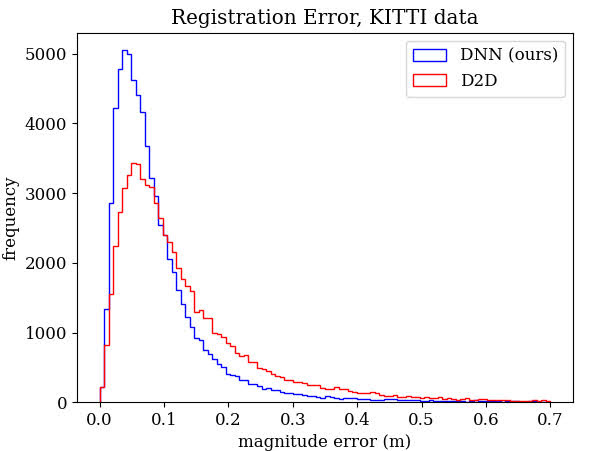}
\caption{D2D and DNN results produce similar levels of error on uniformly sampled ModelNet40 data (left). DNN achieves significantly higher accuracy than D2D on shadowed ModelNet40 point clouds (center). DNN acheives higher accuracy than D2D on KITTI data (right)}
\label{fig:Experiment1Results}
\end{figure*}

\begin{figure*}[h]
\centering
\includegraphics[width=1.9in]{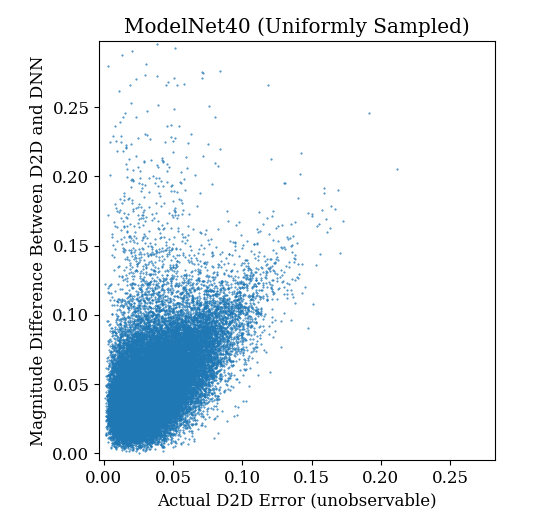}
\includegraphics[width=1.8in]{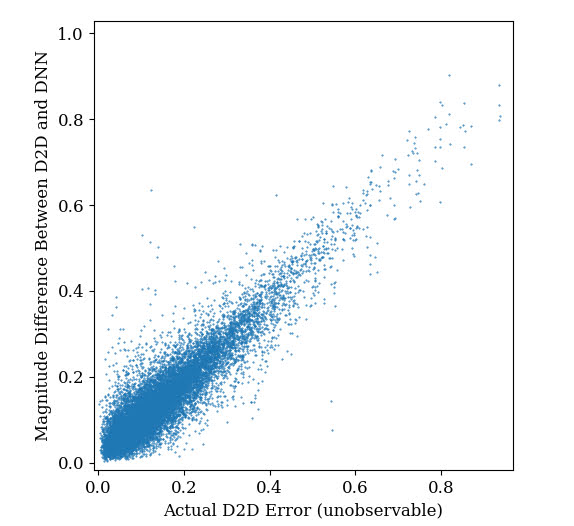}
\includegraphics[width=1.8in]{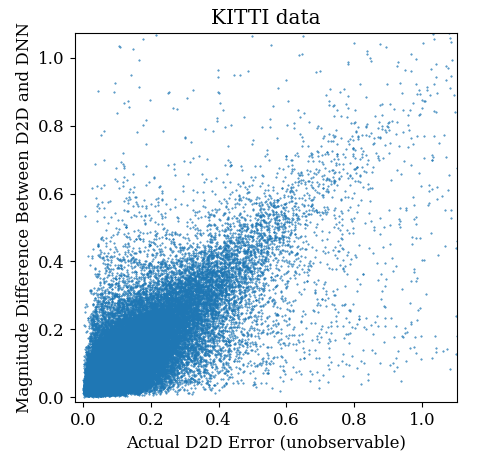}
\caption{Correlation between Magnitude Difference between D2D/DNN estimates and unobservable error of D2D estimate for uniformly sampled ModelNet40 data (left), simulated LIDAR sampled ModelNet40 data (center), and real LIDAR data from KITTI dataset (right)}
\label{fig:covarianceOfMonitor}
\end{figure*}

\begin{table}[h]
    \caption{Experiment II: RMS Forward Translation Error (cm)} 
    \setlength{\tabcolsep}{0.3\tabcolsep}
    \begin{tabularx}{0.48\textwidth} {  
      | >{\centering\arraybackslash}X
      | >{\centering\arraybackslash}X
      | >{\centering\arraybackslash}X 
      | >{\centering\arraybackslash}X
      | >{\centering\arraybackslash}X
      | >{\centering\arraybackslash}X
      | >{\centering\arraybackslash}X
      | >{\centering\arraybackslash}X
      | >{\centering\arraybackslash}X | }
    \hline
     & no voxel rejection & 2$\sigma$ rejection only & 2$\sigma$ \& PointNet rejection (ours) \\
    \hline
    Scene 1 ~(city) & 1.47 & 1.32 & 1.34 \\
    \hline
    Scene 2 (forest) & 5.09 & 4.63 & 4.45 \\ 
    \hline
    Scene 3 (residential) & 3.70 & 2.75 & 2.60 \\ 
    \hline
    \end{tabularx}
    \label{tab:experiment2}
\end{table}

The first trial of Experiment I compared the two methods on uniformly sampled point clouds generated from ModelNet40 objects. In this trial, the D2D method outperformed the DNN solution. Interestingly however, the DNN achieved almost the same accuracy in the simulated LIDAR data as in the uniform sampling, while the D2D estimate performed significantly worse when attempting to register partial point clouds. In the third trial, our DNN outperformed the D2D again, though by a slightly smaller margin.

In Experiment II, the city, forest, and residential trials all used the same network and NDT parameters. For each scene, we compared root-mean-square error of the global NDT registration against a GNSS/INS baseline. Our first test case consists of the raw NDT registration with no rejected voxels. We then consider the commonly employed strategy of removing any voxel with a local D2D residual greater than two standard deviations from the mean residual distance after a preliminary convergence, as prescribed by \cite{magnussonThesis}. 
Finally, we consider the case where our filter is used in conjunction with the 2$\sigma$ rejection criteria. For the city test scene, our proposed method achieved slightly worse accuracy than the 2$\sigma$-only rejection criteria by roughly 1\%. The raw baseline does worse than either, presumably because moving objects are preventing the registration algorithm from converging on the correct solution. In the forest scene, however, our PointNet + 2$\sigma$ combined approach does better than the 2$\sigma$-only method by a margin of roughly 4\%. Similarly, our DNN based approach outperforms the 2$\sigma$-only rejection method in the residential dataset by a margin of over 5.5\%.
Furthermore, because the errors reported in Table (\ref{tab:experiment2}) are relative to an imperfect baseline 
the true odometry error is likely lower in all cases, making the true percentage reduction in error from our proposed method even more significant. 







\section{Discussion}

The results in Table (\ref{tab:experiment1}) and Fig. (\ref{fig:Experiment1Results}) demonstrate our PointNet-based registration network achieves significantly better accuracy than a simple D2D registration on realistic point cloud data, but slightly worse than the D2D strategy on the unrealistic uniformly sampled \textit{ModelNet40} data. The D2D method performed very well in the first trial due to the high degree of symmetry in the data, but performed far worse on the subsequent trials. Our DNN performance, on the other hand, remained relatively insensitive to using full or partial point clouds, achieving similar accuracy between trials 1 and 2. This is likely because the feature vector formed by our PointNet can infer the contours of an object without requiring total overlap between the two scans.

\begin{figure}[h]
\centering
\includegraphics[width=3.4in]{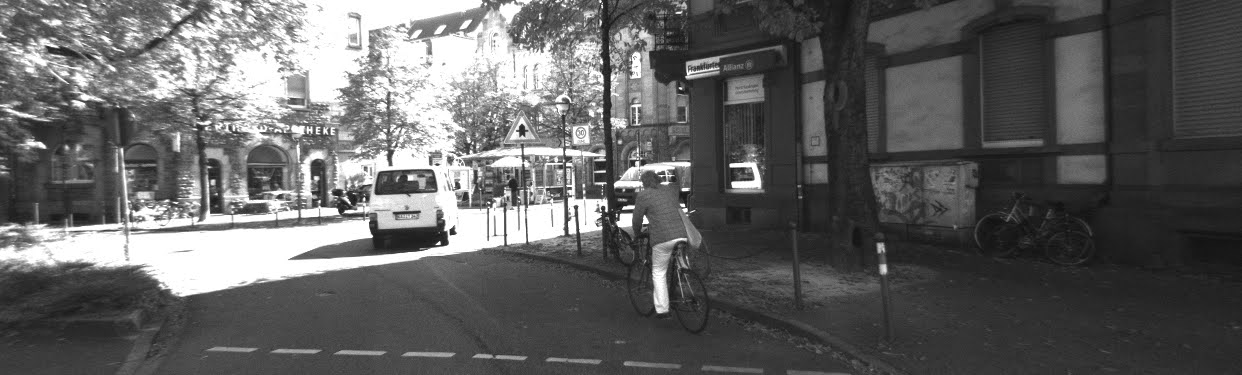}
\includegraphics[width=3.4in]{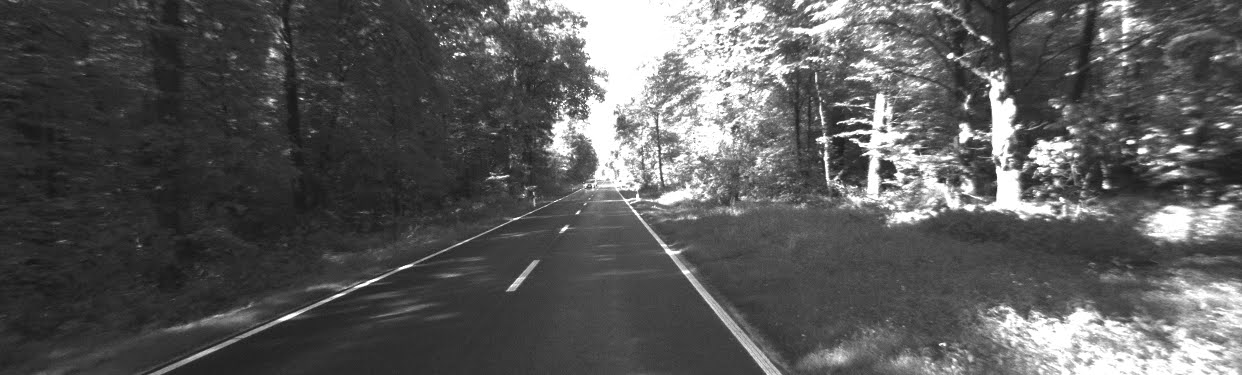}
\includegraphics[width=3.4in]{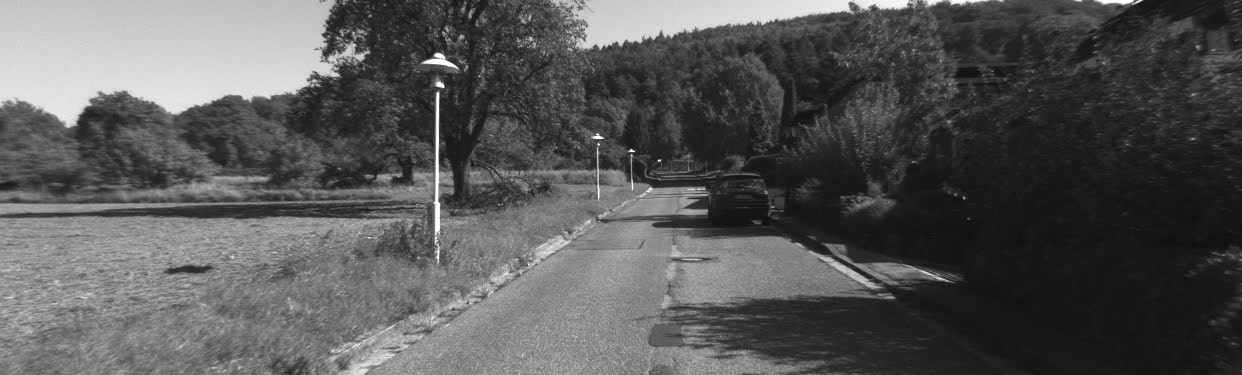}
\caption{KITTI Raw 0005 (top), KITTI Raw 0027 (center), and KITTI Odometry 03 (bottom) }
\label{fig:KITTI1vs2}
\end{figure}


The results in Table (\ref{tab:experiment2}) demonstrate that performance enhancements provided by our DNN filter are highly dependant on the driving environment. For the urban case, there is little benefit to removing voxels with our network, while the forest and residential scenes achieve substantial benefit from our proposed technique. This is possibly due to the fact that in urban settings, the influence of self-occlusion prone surfaces on constraining translation in the forward direction is dwarfed by other large, highly-localizable surfaces, such as the sides of buildings. The 2$\sigma$ rejection method does a good job of identifying moving objects when surfaces are well defined (such as in the city trial) because the global standard deviation of residuals at convergence is small. In more ill-defined situations (such as those containing lots of dense foliage), the standard deviation of residual error at a converged solution is so large that it is unable to reject obviously biased voxels. 

\begin{figure}[h]
\centering
\includegraphics[width=3.2in]{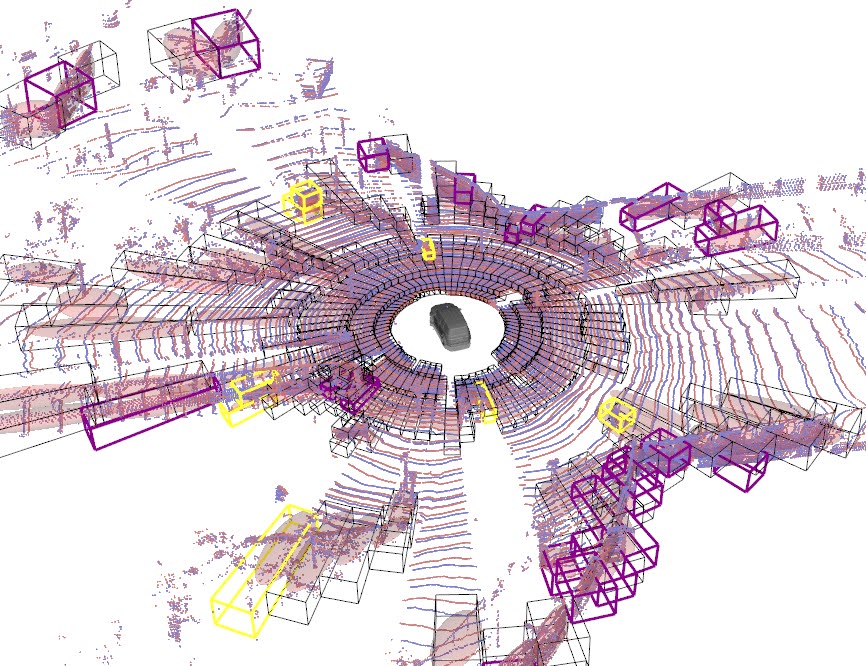}
\includegraphics[width=3.2in]{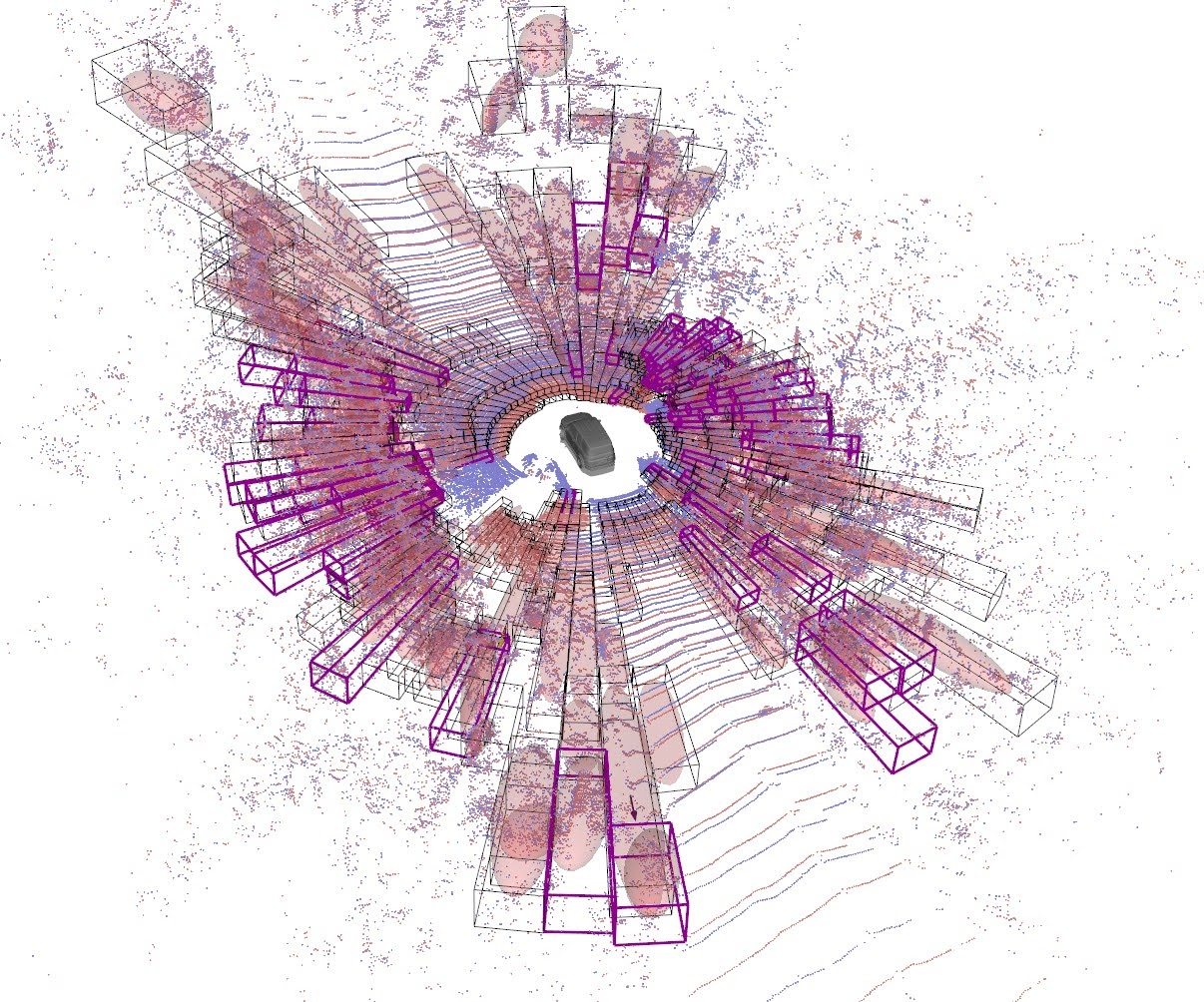}
\caption{Sequential point clouds from the city (top) and forest (bottom) trajectories registered by our algorithm. Occupied spherical voxels drawn containing distributions from the keyframe scan. Voxels identified as containing moving objects (via the 2$\sigma$ rejection filter) are highlighted in yellow. Voxels flagged as biased by our DNN filter are highlighted in purple.}
\label{fig:KITTIoverhead}
\end{figure}

The scanning noise normal to surfaces (either due to time of flight error from the sensor or surface roughness) also plays an important role in the performance benefits of our proposed bias mitigation strategy. In our testing, we noted that the "fuzzier" the surfaces, the more beneficial our bias mitigation procedure became. Recall, D2D registration techniques use the mahalanobis distance to weigh residuals between corresponding distribution ellipsoids \cite{D2DNDT}. Scenes with higher surface normal noise lead to situations where flat surfaces (such the side of the building in Fig. (\ref{fig:KITTI1vs2}) are not oblate enough to sufficiently resist the influence of high perspective shift bias in other regions of the frame. The surfaces that tend to produce the highest perspective shift error are either concave or convex, and thus generate distribution ellipsoids that are far less oblate (and therefore weighted less) than their well conditioned counterparts, by virtue of a  gaussian's poor ability to characterize a non-planar surfaces. When overall noise is low, these features are heavily deweighted. On the other hand, higher overall noise effectively raises the lower bound on compactness for any distribution of points, thereby deweighting the contributions of flat, well-conditioned surfaces and allowing the biased portions of the scan to dominate registration. 

Bias from self-occlusion is also highly dependant on vehicle speed. This is because sequential point clouds are recorded from further away from one another as the vehicle speeds up. At large distances between frames, small objects close to the vehicle are observed from different angles-- capturing different faces of the same object violates the assumption of a static scene and can cause the voxel to incorrectly attempt to register non-corresponding parts of the object. This effect is particularly exaggerated in the forest experiment.

Fig. (\ref{fig:KITTIoverhead}) presents a visualization of two pairs of point clouds and highlights the voxels flagged as potentially biased. In the forest scene, the flagged voxels are primarily aligned close to the sides of the vehicle and normal to the direction of travel. This is where perspective shift bias is most severe-- the fact that these voxels are being suppressed explains the performance benefits of our network in forested scenes.
In the city scene, however, the voxels flagged for removal are fewer in number and spread much more uniformly throughout the frame. It is likely a large number of these identified voxels are false positives rather than systemic error so removing them from the registration estimate ignores a small amount of useful information, which explains the slight decrease in solution accuracy. Trading solution accuracy in well structured environments for solution integrity in unstructured and novel environments is useful for safety critical applications. Furthermore, future work in increasing the accuracy of the registration network can decrease the number of erroneously removed voxels.   

The implementation demonstrated in this paper makes use of a simple registration network trained on a relatively small dataset using a single NVIDIA RTX 3090. Our network structure only considers the cluster of points inside each voxel, ignoring any potential queues from their context in the larger frame. Despite these limitations, our method manages to reduce error in D2D scan registrations in unstructured forest and residential scenes, and does not significantly impact accuracy when applied in a well-structured city environment. Future work in enhancing network structure, especially aggregating contextual information between voxels has the potential to further increase overall registration accuracy.
\section{Conclusion}

In this paper we discussed a new method for reducing bias in Distribution-to-Distribution scan registration. Our proposed technique removes portions of LIDAR scans that violate the assumption of a static scene by identifying and discarding voxels that locally disagree with a PointNet-based registration network. Our method achieves results similar to existing classical outlier rejection techniques on well structured city scenes and outperforms existing techniques in unstructured forest and residential environments. Compared to other machine learning-based bias mitigation techniques, our strategy is unique in that it does not directly provide a correction factor, and instead merely removes data from the classical optimization routine that is likely to introduce error. This allows the registration algorithms to adjust estimates for solution error covariance as data is removed from the frame, and allows the scan matching process to remain fully interpretable.

{\small
\bibliographystyle{ieee_fullname}
\bibliography{references}
}

\end{document}